\documentclass[11pt,a4paper,usenames]{article}
\usepackage{emnlp2018}
\usepackage{times}
\usepackage{latexsym,graphicx,amssymb,multirow,amsmath,booktabs}
\usepackage{soul}
\usepackage{subfigure}
\usepackage[font={footnotesize}]{caption}
\usepackage{url}
\usepackage{algorithm}
\usepackage{algpseudocode}
\usepackage{balance}
\newcommand{\ModelName}{{\textsc{IntentCapsNet-ZSL}}}
\newcommand{\IntentModelName}{{\textsc{IntentCapsNet}}}
\newcommand{\UnseenCapsule}{Zero-shot DetectionCaps}
\newcommand{\SecondCapsule}{DetectionCaps}
\aclfinalcopy % Uncomment this line for the final submission

\pdfoutput=1
\newcommand{\hlcy}[2][yellow]{{%
    \colorlet{foo}{blue!#1}%
    \sethlcolor{foo}\hl{#2}}}

\newcommand{\hlcr}[2][yellow]{{%
    \colorlet{foo}{blue!#1}%
    \sethlcolor{foo}\hl{#2}}}

\title{Zero-shot User Intent Detection via Capsule Neural Networks}

\author{Congying Xia$^1$\thanks{~ Indicates Equal Contribution. \newline Previously avilable on \url{http://doi.org/10.13140/RG.2.2.11739.46889}},~Chenwei Zhang{$^1$\footnotemark[1]}, ~Xiaohui Yan$^2$, ~Yi Chang$^{3,4,5}$, ~Philip S. Yu$^1$\\
  {$^1$Department of Computer Science, University of Illinois at Chicago, Chicago, IL 60607 USA}\\
  {$^2$Huawei Technologies, San Jose, CA 95050 USA}\\
  {$^3$College of Artificial Intelligence, Jilin University, Changchun, China}\\
  {$^4$College of Computer Science and Technology, Jilin University, Changchun, China}\\
  {$^5$Key Laboratory of Symbolic Computation and Knowledge Engineering of Ministry of Education, China}\\
  {\tt \{cxia8,czhang99,psyu\}@uic.edu},
  {\tt yanxiaohui2@huawei.com},
  {\tt yichang@acm.org}\\
}

\begin{document}
\maketitle
\begin{abstract}
User intent detection plays a critical role in question-answering and dialog systems. Most previous works treat intent detection as a classification problem where utterances are labeled with predefined intents. However, it is labor-intensive and time-consuming to label users' utterances as intents are diversely expressed and novel intents will continually be involved. Instead, we study the zero-shot intent detection problem, which aims to detect emerging user intents where no labeled utterances are currently available. We propose two capsule-based architectures: {\IntentModelName} that extracts semantic features from utterances and aggregates them to discriminate existing intents, and {\ModelName} which gives {\IntentModelName} the zero-shot learning ability to discriminate emerging intents via knowledge transfer from existing intents.
Experiments on two real-world datasets show that our model not only can better discriminate diversely expressed existing intents, but is also able to discriminate emerging intents when no labeled utterances are available. 
\end{abstract}

\section{Introduction}
With the increasing complexity and accuracy of speech recognition technology, companies are striving to deliver intelligent conversation understanding systems as people interact with software agents that run on speaker devices or smart phones via natural language interface \citep{hoy2018alexa}. 
Products like Apple's Siri, Amazon's Alexa and Google Assistant are able to interpret human speech and respond them via synthesized voices. 

\begin{figure*}[h]
    \centering
    \includegraphics[width=15cm]{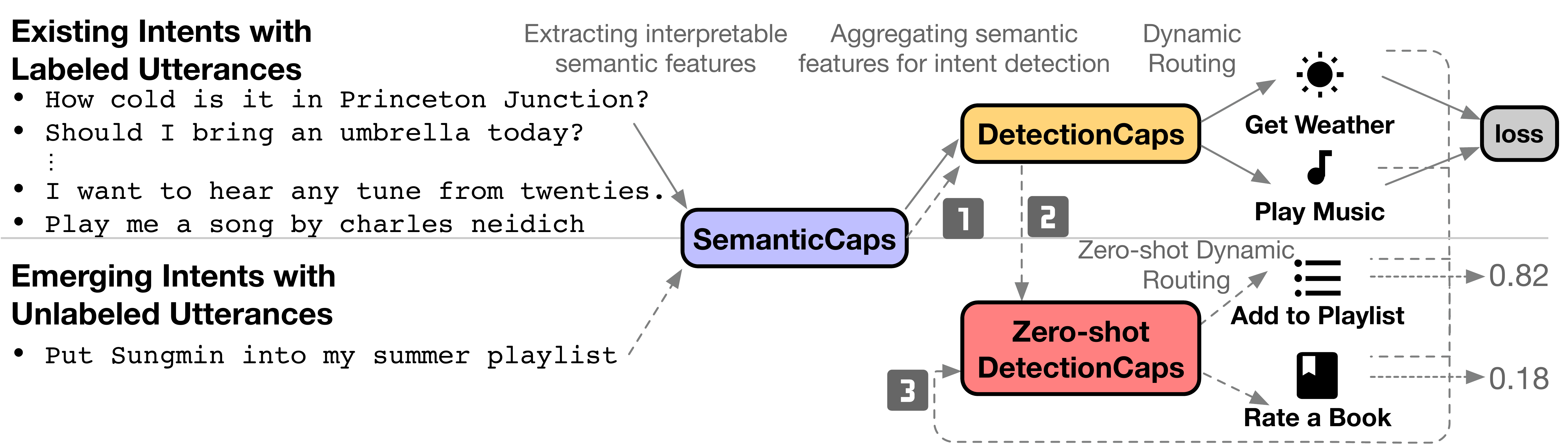}
    \vspace{0.1in}
    \caption{Illustration of the proposed {\ModelName} model for zero-shot intent detection: labeled utterances with existing intents like \textsf{GetWeather} and \textsf{PlayMusic} are used to train an intent detection classifier among existing intents, in which SemanticCaps extract intepretable semantic features and DetectionCaps dynamically aggregate semantic features for intent detection using a novel routing-by-agreement mechanism. For emerging intents, {\ModelName} builds zero-shot DetectionCaps that utilize the (1) outputs of SemanticCaps, (2) the routing information on existing intents from DetectionCaps, and (3) similarities of the emerging intent label to existing intent labels to discriminate emerging intents like \textsf{AddToPlayist} from \textsf{RateABook}. Solid lines indicate the training process and dash lines indicate the zero-shot inference process. 
    }
    \label{fig:overall}
\end{figure*}

With recent developments in deep neural networks, user intent detection models \citep{hu2009understanding,xu2013convolutional,zhang2016mining,liu2016attention,chen2016end} are proposed to classify user intents given their diversely expressed utterances in the natural language. The decent performances on intent detection usually come with deep neural network classifiers optimized on large-scale utterances which are human-labeled among existing predefined user intents.

As more features and skills are being added to devices which expand their capabilities to new programs, it is common for voice assistants to encounter the scenario where no labeled utterance of an emerging user intent is available in the training data, as illustrated in Figure \ref{fig:overall}.
Current intent detection methods train classifiers in a supervised fashion and they are good at discriminating existing intents such as \textsf{Get Weather} and \textsf{Play Music} whose labeled utterances are already available.
However, these models, by the nature of designs, are incapable to detect utterances of emerging intents 
like \textsf{AddToPlaylist} and \textsf{RateABook},
since no labeled utterances are available.
Moreover, it's labor-intensive and time-consuming to annotate utterances of emerging intents and retrain the whole intent detection model. 

Thus, it is imperative to develop intent detection models with the zero-shot learning (ZSL) ability \citep{lampert2014attribute, socher2013zero, changpinyo2016synthesized}: the ability to expand classifiers and the intent detection space beyond the existing intents, of which we have labeled utterances during training, to emerging intents, of which no labeled utterances are available.

The research on zero-shot intent detection is still in its infancy. Previous zero-shot learning methods for intent detection utilize external resources such as label ontologies \citep{ferreira2015online, ferreira2015zero} or manually defined attributes that describe intents \citep{yazdani2015model} to associate existing and emerging intents, which require extra annotation.
Compatibility-based methods for zero-shot intent detection \citep{chen2016zero,kumar2017zero} assume the capability of learning a high-quality mapping from the utterance to its intent directly, so that such mapping can be further capitalized to measure the compatibility of an utterance with emerging intents. However, the diverse semantic expressions may impede the learning of such mapping.

In this work, we make the very first attempt to tackle the zero-shot intent detection problem with a capsule-based \citep{hinton2011transforming,sabour2017dynamic} model.
A capsule houses a vector representation of a group of neurons, and the orientation of the vector encodes properties of an object (like the shape/color of a face), while the length of the vector reflects its probability of existence (how likely a face with certain properties exists). 
The capsule model learns a hierarchy of feature detectors via a routing-by-agreement mechanism: capsules for detecting low-level features (like nose/eyes) send their outputs to high-level capsules (such as faces) only when there is a strong agreement of their predictions to high-level capsules.

The aforementioned properties of capsule models could be quite appealing for text modeling, specifically in this case, modeling the user utterance for intent detection: 
low-level semantic features such as the get\_action, time and city\_name contribute to a more abstract intent (\textsf{GetWeather}) collectively. A semantic feature, which may be expressed quite differently among users, can contribute more to one intent than others. The dynamic routing-by-agreement mechanism can be used to dynamically assign a proper contribution of each semantic and aggregate them to get an intent representation.

More importantly, we discover the potential of zero-shot learning ability on the capsule model, which is not yet widely recognized. It makes the capsule model even more suitable for text modeling when no labeled utterances are available for emerging intents. 
The ability to neglect the disagreed output of low-level semantics for certain intents during routing-by-agreement encourages the learning of generalizable semantic features that can be adapted to emerging intents. 
For each emerging intent with no labeled utterances, a {\UnseenCapsule} is constructed explicitly by using not only semantic features SemanticCaps extracted, but also existing routing agreements from {\SecondCapsule} and similarities of an emerging intent label to existing intent labels.

In summary, the contributions of this work are: 

$\bullet$ Expanding capsule neural networks to text modeling, by extracting and aggregating semantics from utterances in a hierarchical manner; 

$\bullet$ Proposing a novel and effective capsule-based model for zero-shot intent detection;

$\bullet$ Showing and interpreting the effectiveness of our model on two real-world datasets.
\section{Problem Formulation}
In this section, we first define related concepts, and formally state the problem.

\noindent\textbf{Intent.}
An intent is a purpose, or a goal that underlies a user-generated utterance \citep{ibm2017doc}. 
An utterance can be associated with one or multiple intents. 
We only consider the basic case that an utterance is with a single intent.
However, utterances with multiple intents can be handled by segmenting them into single-intent snippets using sequential tagging tools like CRF \citep{lafferty2001conditional}, which we leave for future works. 

\noindent\textbf{Intent Detection.}
Given a labeled training dataset where each sample has the following format: \((x, y)\) where $x$ is an utterance and $y$ is its intent label, each training example is associated with one of $K$ existing intents $y \in{Y}=\{y_1, y_2, ... , y_K\}$. The intent detection task tries to associate an utterance $x_{existing}$ with its correct intent category in the existing intent classes $Y$.

\noindent\textbf{Zero-shot Intent Detection.}
Given the labeled training set \{$(x,y)$\} where $y{\in}Y$, 
the zero-shot intent detection task aims to detect an utterance $x_{emerging}$ which belongs to one of $L$ emerging intents $z{\in}Z=\{z_1, z_2, ... , z_L\}$ where $Y{\cap}Z=\varnothing$.
\section{Approach}
We propose two architectures based on capsule models: {\IntentModelName} that is trained to discriminate among utterances with existing labels, e.g. existing intents for intent detection; {\ModelName} that gives zero-shot learning ability to {\IntentModelName} for discriminating unseen labels, i.e. emerging intents in this case.
As shown in Figure \ref{fig:train_test}, the cores of the proposed architectures are three types of capsules: SemanticCaps that extract interpretable semantic features from the utterance, {\SecondCapsule} that aggregate semantic features for intent detection, and {\UnseenCapsule} which discriminate emerging intents.

\iftrue
\begin{figure}[tbh!]
    \centering
    \includegraphics[width=\linewidth]{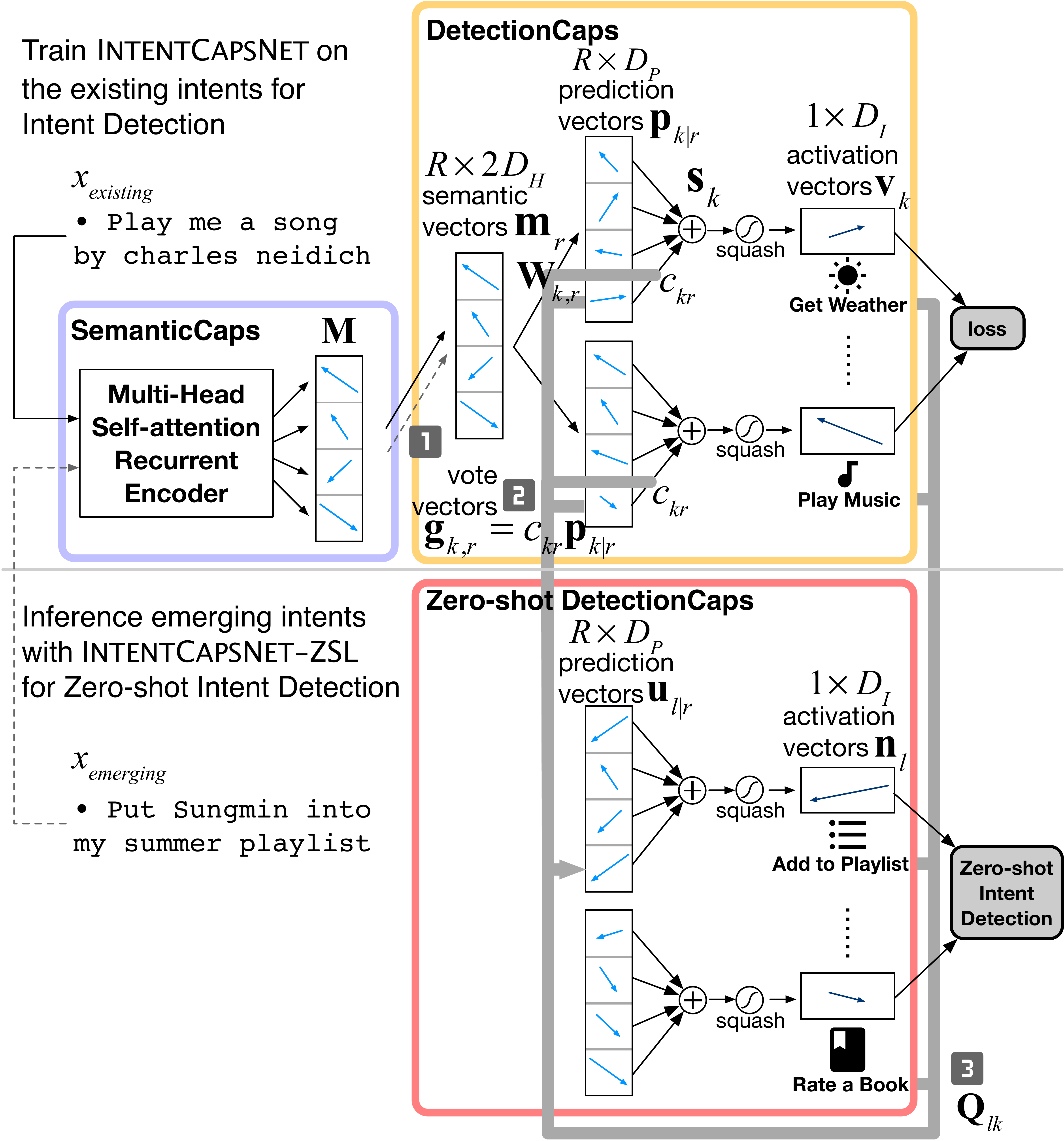}
    \vspace{0.1in}
    \caption{
    The architecture of {\IntentModelName} and {\ModelName}. During training, utterances with existing intents are fed into the SemanticCaps which output vectorized semantic features, i.e. semantic vectors. Then {\SecondCapsule} combine these features into higher-level prediction vectors and output an activation vector for intent detection on each existing intent. During inference, emerging utterances take advantages of the SemanticCaps trained in {\IntentModelName} to extract semantic features from the utterance (shown in 1), then the vote vectors on the existing intents are transferred to emerging intents (shown in 2) using similarities between existing and emerging intents (shown in 3). The obtained activation vectors for emerging intents are used for zero-shot intent detection.
    }
    \label{fig:train_test}
\end{figure}
\fi

\subsection{SemanticCaps}
In the original capsule model \citep{sabour2017dynamic}, convolution-based PrimaryCaps are introduced as the first layer to obtain different vectorized features from the raw input image. While in this work, an intrinsically similar motivation is adopted to extract different semantic features from the raw utterance by a new type of capsule named SemanticCaps. Unlike the PrimaryCaps which use convolution operators with a large reception field to extract spacial-proximate features, the SemanticCaps is based on a bi-direction recurrent neural network with multiple self-attention heads, where each self-attention head focuses on certain part of the utterance and extracts a semantic feature that may not be expressed by words in proximity. 

Given an input utterance \(x = \left (  \mathbf{w_{1}}, \mathbf{w_{2}}, ..., \mathbf{w_{T}}\right )\) of $T$ words, each word is represented by a vector of dimension $D_W$ that can be pre-trained using a skip-gram language model \cite{mikolov2013efficient}. A recurrent neural network such as a bidirectional LSTM \citep{hochreiter1997long} is applied to sequentially encode the utterance into hidden states:
\begin{equation}
\begin{gathered}
    {\mathord{\buildrel{\lower3pt\hbox{$\scriptscriptstyle\rightarrow$}}\over {\mathbf{h}}} }_{t} = {\rm LSTM}_{fw} (\mathbf{w}_{t}, {\mathord{\buildrel{\lower3pt\hbox{$\scriptscriptstyle\leftarrow$}}\over {\mathbf{h}}} }_{t-1}),\\
     {\mathord{\buildrel{\lower3pt\hbox{$\scriptscriptstyle\leftarrow$}}\over {\mathbf{h}}} }_{t} = {\rm LSTM}_{bw} (\mathbf{w}_{t}, {\mathord{\buildrel{\lower3pt\hbox{$\scriptscriptstyle\leftarrow$}}\over {\mathbf{h}}} }_{t+1}).\\
\end{gathered}
\end{equation}
For each word \(\mathbf{w_t}\), we concatenate each forward hidden state \(\vec{\mathbf{h}}_{t}\) obtained from the forward ${\rm LSTM}_{fw}$ with a backward hidden state \({\mathord{\buildrel{\lower3pt\hbox{$\scriptscriptstyle\leftarrow$}}\over 
{\mathbf{h}}} }_{t}\) from ${\rm LSTM}_{bw}$ to obtain a hidden state \(\mathbf{h_t}\) for the word \(\mathbf{w_t}\). The whole hidden state matrix can be defined as \(\mathbf{H} = \left(\mathbf{h}_{1}, \mathbf{h}_{2}, ..., \mathbf{h}_{T}\right) \in \mathbb{R}^{T \times 2D_{H}}\), where $D_H$ is the number of hidden units in each LSTM.

Inspired by the success of self-attention mechanisms \citep{vaswani2017attention,lin2017structured} 
for sentence embedding, 
we adopt a multi-head self-attention framework where each self-attention head is encouraged to be attentive to a specific semantic feature of the utterance, such as certain sets of keywords or phrases in the utterance: one self-attention may be attentive for the ``get'' action in \textsf{GetWeather}, while another one may be attentive to city\_name in \textsf{GetWeather}: it decides for itself what semantics to be attentive to.

A self-attention weight matrix $\mathbf{A}$ is computed as:
\begin{equation}
\mathbf{A} = {\rm softmax}\left ( \mathbf{W}_{s2} {\rm tanh}\left (  \mathbf{W}_{s1}\mathbf{H}^{T} \right ) \right ),
\end{equation}
where \( \mathbf{W}_{s1} \in \mathbb{R}^{D_{A} \times 2D_{H}}\) and \( \mathbf{W}_{s2} \in \mathbb{R}^{ R \times D_{A}}\) are weight matrices for the self-attention.
\(D_{A}\) is the hidden unit number of self-attention and \(R\) is the number of self-attention heads. The softmax function makes sure for each self-attention head, the attentive scores on all the words sum to one.

A total number of $R$ semantic features are extracted from the input utterance, each from a separate self-attention head:
$
     \mathbf{M} =  \mathbf{A}\mathbf{H},
$
where \(\mathbf{M} =  \left( \mathbf{m}_{1}, \mathbf{m}_{2}, ..., \mathbf{m}_{R}\right) \in \mathbb{R}^{R \times 2D_{H}}\). Each $\mathbf{m}_r$ is a $2D_{H}-$dimensional semantic vector.

Each semantic vector will have a distinguishable orientation when the objective is properly regularized (details in Equation 6), as we want each attention to be attentive to a unique semantic feature of the utterance.
The vector representation adopted in capsules is suitable to portray the low-level semantic properties as well as high-level intents of the utterance, where the orientation of a vector represents semantic/intent properties that may slightly vary depending on the expressions. 
The capsule encourages the learning of generalizable semantic vectors: less informative semantic properties for one intent may not be penalized by their orientations: they simply possess small norms as they are less likely to exist.

\subsection{\SecondCapsule}
The output of SemanticCaps are low-level vector representations of $R$ different semantic features extracted from the utterances.
To combine these features into higher-level representations,
we build {\SecondCapsule} that choose different semantic features dynamically so as to form an intent representation for each intent via an unsupervised routing-by-agreement mechanism.

As a semantic feature may contribute differently in detecting different intents, the {\SecondCapsule} first encode semantic features with respect to each intent:
\begin{equation}
    \mathbf{p}_{k|r} = \mathbf{m}_{r}\mathbf{W}_{k,r},
\end{equation}
where $k\in\{1, 2, ..., K\}$, $r\in\{1, 2, ..., R\}$. $\mathbf{W}_{k,r}\in \mathbb{R}^{2D_H\times{D_P}}$ is the weight matrix of the {\SecondCapsule},
\(\mathbf{p}_{k|r}\) is the prediction vector of the $r$-th semantic feature of an existing intent \(k\),
and ${D_P}$ is the dimension of the prediction vector.

\noindent\textbf{Dynamic Routing-by-agreement.}
The prediction vectors obtained from SemanticCaps route dynamically to {\SecondCapsule}.
The {\SecondCapsule} computes a weighted sum over all prediction vectors:
\begin{equation}
    \mathbf{s}_{k} = \sum_{r}^{R}c_{kr}\mathbf{p}_{k|r},
\end{equation}
where \(c_{kr}\) is the coupling coefficient that determines how informative, or how much contribution the $r$-th semantic feature is to the intent $y_k$.  \(c_{kr}\) is calculated by an unsupervised, iterative dynamic routing-by-agreement algorithm \citep{sabour2017dynamic}, which is briefly recalled in Algorithm \ref{al:routing}. As shown in this algorithm, ${b}_{kr}$ is the initial logit representing the log prior probability that a SemanticCap r is coupled to an DetectionCap k.
\begin{algorithm}[h]
\caption{Dynamic routing algorithm}
\label{al:routing}
\resizebox{7cm}{!}{% <-
\begin{minipage}{1.15\linewidth}
\begin{algorithmic}[1]
\Procedure{Dynamic Routing}{$\mathbf{p}_{k|r}$, $iter$}
    \State {for all semantic capsule r and intent capsule k: ${b}_{kr}\leftarrow0$.}
    
    \For {$iter$~iterations}
        \State {for all SemanticCaps r: $\mathbf{c}_r\leftarrow\operatorname{softmax}(\mathbf{b}_r)$}
        
        \State {for all {\SecondCapsule} k: $\mathbf{s}_k \leftarrow \Sigma_r c_{kr}\mathbf{p}_{k|r}$}
        
        \State {for all {\SecondCapsule} k: $\mathbf{v}_k= \operatorname{squash}(\mathbf{s}_{k})$}
        
        \State {for all SemanticCaps r and {\SecondCapsule} k: \phantom{0000000000}${b}_{kr} \leftarrow \text{b}_{kr} + \mathbf{p}_{k|r}\cdot\mathbf{v}_{k}$}
        
	    \EndFor
	\State Return $\mathbf{v}_k$
\EndProcedure
\end{algorithmic}
\end{minipage}}
\end{algorithm}
 
The squashing function $\operatorname{squash}(\cdot)$ is applied on \(\mathbf{s}_{k}\) to get an activation vector \(\mathbf{v}_{k}\) for each existing intent class \(k\):
\begin{equation}
    \mathbf{v}_{k} = \frac{\left \| \mathbf{s}_{k} \right \|^{2}}{1 + \left \| \mathbf{s}_{k}\right \|^{2} }\frac{\mathbf{s}_{k}}{\left \| \mathbf{s}_{k} \right \|},
\end{equation}
where the orientation of the activation vector $\mathbf{v}_k$ represents intent properties while its norm indicates the activation probability. 
The dynamic routing-by-agreement mechanism assigns low $c_{kr}$ when there is inconsistency between $p_{k|r} $ and $v_k$, which ensures the outputs of the SemanticCaps get sent to appropriate subsequent {\SecondCapsule}.

\noindent\textbf{Max-margin Loss for Existing Intents.}
The loss function considers both the max-margin loss on each labeled utterance, as well as a regularization term that encourages each self-attention head to be attentive to a different semantic feature of the utterance:
\begin{align}\label{eq:loss}
\begin{split}
\mathcal{L}&= \sum_{k=1}^{K}\{
\left[\kern-0.15em\left[ {{y} = y_k} \right]\kern-0.15em\right]\cdot \max (0,m^ +  - {{\left\| {{\mathbf{v}_k}} \right\|}})^2\\
&+ \lambda \left[\kern-0.15em\left[ {{y} \ne y_k} \right]\kern-0.15em\right]\cdot \max (0,{{\left\| {{\mathbf{v}_k}} \right\|}} - m^ - )^2\}\\
&+ \alpha||{\textbf{A}}\textbf{A}^{T}-I||^2_F,
\end{split}
\end{align}
where $\left[\kern-0.15em\left[  \right]\kern-0.15em\right]$ is an indicator function, $y$ is the ground truth intent label for the utterance $x$, $\lambda$ is a down-weighting coefficient, $m^ +$ and $m^ -$ are margins. $\alpha$ is a non-negative trade-off coefficient that encourages the discrepancies among different attention heads.

\subsection{\UnseenCapsule}
To detect emerging intents effectively, {\UnseenCapsule} are designed to transfer knowledge from existing intents to emerging intents.

\noindent\textbf{Knowledge Transfer Strategies.} 
As SemanticCaps are trained to extract semantic features from utterances with various existing intents, a self-attention head which has similar extraction behavior among existing and emerging intents may help transfer knowledge. For example, a self-attention head that extracts the ``play'' action mentioned by \texttt{turn on/I want to hear} in the beginning of an utterance for \textsf{PlayMusic} is helpful if it is also attentive to expressions for the ``add'' action like \texttt{add/I want to have} in the beginning of an utterance with an emerging intent \textsf{AddtoPlaylist}.

The coupling coefficient $c_{kr}$ learned by  {\SecondCapsule} in a totally unsupervised fashion embodies rich knowledge of how informative $r$-th semantic is to the existing intent $k$. We can capitalize on the existing routing information for emerging intents. For example, how the word \texttt{play} routes to \textsf{GetWeather} can be helpful in routing the word \texttt{add} to \textsf{AddtoPlaylist}.

The intent labels also contain knowledge of how two intents are similar with each other. For example, an emerging intent \textsf{AddtoPlaylist} can be closer to one existing intent \textsf{PlayMusic} than \textsf{GetWeather} due to the proximity of the embedding of \textsf{Playlist} to \textsf{Play} or \textsf{Music}, than \textsf{Weather}.
\vspace{-0.02in}

\noindent\textbf{Build Vote Vectors.}
As the routing information and the semantic extraction behavior are strongly coupled ($c_{kr}$ is calculated by $\mathbf{p}_{k|r}$ iteratively in Line 4-6 of Algorithm \ref{al:routing}) and their products are summarized to get the activation vector $v_k$ for intent $k$ (Line 5-6 of Algorithm \ref{al:routing}), we denote vectors before summation as vote vectors:
%\vspace{-0.1in}
\begin{equation}
    \textbf{g}_{k,r} = c_{kr}\textbf{p}_{k|r},
    %\vspace{-0.1in}
\end{equation}
%\vspace{-0.02in}
where \(\textbf{g}_{k, r}\) is the \(r\)-{th} vote vector for an existing intent \(k\).

\noindent\textbf{Zero-shot Dynamic Routing.}
The zero-shot dynamic routing utilizes vote vectors from existing intents to build intent representations for emerging intents via a similarity metric between existing intents and emerging intents.

Since there are \(K\) existing intents and \(L\) emerging intents, the similarities between existing and emerging intents form a matrix \(\mathbf{Q} {\in} \mathbb{R}^{L \times K}\).
Specifically, the similarity between an emerging intent \(z_l {\in} Z\) and an existing intent \(y_k {\in} Y\) is computed as:
\begin{equation}
    q_{lk} = \frac{exp\left \{ -d\left ( \mathbf{e}_{z_l}, \mathbf{e}_{y_k} \right ) \right \}}{\sum_{k=1}^{K}exp\left \{ -d\left ( \mathbf{e}_{z_l}, \mathbf{e}_{y_k} \right ) \right \}},
\end{equation}
where 
\begin{equation}
    d\left(\mathbf{e}_{z_l}, \mathbf{e}_{y_k}\right) = \left(\mathbf{e}_{z_l}-\mathbf{e}_{y_k}\right)^{T}\Sigma^{-1}\left(\mathbf{e}_{z_l}-\mathbf{e}_{y_k}\right).
\end{equation}
\(\mathbf{e}_{z_l}, \mathbf{e}_{y_k} \in \mathbb{R}^{D_{I}\times 1}\) are intent embeddings computed by the sum of word embeddings of the intent label.
\(\Sigma\) models the correlations among intent embedding dimensions and we use \(\Sigma = \sigma ^{2}I\). \(\sigma\) is a hyper-parameter for scaling.
The prediction vectors for emerging intents are thus computed as:
\begin{equation}
    \mathbf{u}_{l|r} = \sum_{k=1}^{K}q_{lk}\mathbf{g}_{k,r}.
\end{equation}
We feed the prediction vector $\mathbf{n}_l$ to Algorithm \ref{al:routing} and derive activation vectors $\mathbf{n}_l$ on emerging intents as the output.
The final intent representation $\mathbf{n}_l$ for each emerging intent is updated toward the direction where it coincides with representative votes vectors.

We can easily classify the utterance of emerging intents by choosing the activation vector with the largest norm ${\hat z} = \mathop {\arg \max }\limits_{z_l \in Z} {\left\| {{{\mathbf{n}}_l}} \right\|}$.
\begin{table*}[h!]
\centering
\resizebox{\textwidth}{!}{
\begin{tabular}{l|cccc|cccc}
\hline
\multirow{2}{*}{\textbf{Model}} & \multicolumn{4}{c|}{\textbf{SNIPS-NLU} (on 5 existing intents)} & \multicolumn{4}{c}{\textbf{CVA} (on 80 existing intents)} \\ \cline{2-9} 
                      & Accuracy &Precision & Recall  & F1
                      & Accuracy &Precision & Recall  & F1              \\ \hline

TFIDF-LR &0.9546 &0.9551 &0.9546 &0.9545 &0.7979	&0.8104	&0.7979	&0.7933\\
TFIDF-SVM &0.9584 &0.9586 &0.9584 &0.9581  &0.7989	&0.8111	&0.7989	&0.7942\\
CNN&0.9595 &0.9596 &0.9595 &0.9595 &0.8223	& 0.8288    &0.8223   & 0.8210\\
RNN &0.9516 &0.9522 &0.9516 &0.9518 &0.8286	&0.8330    &0.8286    &0.8275\\
GRU&0.9535 &0.9535 &0.9535 &0.9534 &0.8239	&0.8281	&0.8239	&0.8216\\
LSTM &0.9569 &0.9573 &0.9569 &0.9569 &0.8319	& 0.8387   & 0.8319    &0.8306\\
Bi-LSTM &0.9501 &0.9502 &0.9501 &0.9502 &0.8428	&0.8479   & 0.8428    &0.8419\\
Self-attention Bi-LSTM &0.9524 &0.9522 &0.9524 &0.9522 &0.8521	&  0.8590    &0.8521    &0.8513\\\hline
\IntentModelName          &\textbf{0.9621}  &\textbf{0.9620}  &\textbf{0.9621}  &\textbf{0.9620} &\textbf{0.9088} &\textbf{0.9160}  &\textbf{0.9088}  &\textbf{0.9023}\\ \hline
\end{tabular}
}
\vspace{0.1in}
\caption{Intention detection results using {\IntentModelName} on two datasets. All the metrics (Accuray, Precision, Recall and F1) are reported using the average value weighted by their support on per class.
}
\label{tab::seen_intents}
\end{table*}
\section{Experiment Setup}
To demonstrate the effectiveness of our proposed models, 
we apply {\IntentModelName} to detect existing intents in an intent detection task, and use {\ModelName} to detect emerging intents in a zero-shot intent detection task.

\noindent\textbf{Datasets.}
For each task, we evaluate our proposed models by applying it on two real-word datasets: SNIPS Natural Language Understanding benchmark (SNIPS-NLU) and a Commercial Voice Assistant (CVA) dataset. The statistical information on two datasets are shown in Table \ref{data_statistics}.
SNIPS-NLU\footnote{https://github.com/snipsco/nlu-benchmark/} is an English natural language corpus collected in a crowdsourced fashion to benchmark the performance of voice assistants.
CVA is a Chinese natural language corpus collected anonymously from a commercial voice assistant on smart phones.

\begin{table}[h!]
\centering
\resizebox{\linewidth}{!}{% <-
\begin{tabular}{l|l|l}
\hline
\textbf{Dataset} & \textbf{SNIPS-NLU} & \textbf{CVA} \\\hline   
Vocab Size   &  10,896 & 1,709\\
Number of Samples   &  13,802    &  9,992  \\
Average Sentence Length   &  9.05    & 4\\
Number of Existing Intents   &  5   & 80\\
Number of Emerging Intents  &  2   & 20\\\hline
\end{tabular}
}
\vspace{0.1in}
\caption{Dataset statistics.}
\label{data_statistics}
\end{table}

\noindent\textbf{Baselines.}
We first compare the proposed capsule-based model {\IntentModelName} with other text classification alternatives on the detection of existing intents:
1) TFIDF-LR/TFIDF-SVM: we use TF-IDF to represent the utterance and use logistic regression/support vector machine as classifiers. 
2) CNN: a convolutional neural network \citep{kim2014convolutional} that uses convolution and pooling operations, which is popular for text classification.
3) RNN/GRU/LSTM/BiLSTM: we adopt different types of recurrent neural networks: the vanilla recurrent neural network (RNN), gated recurrent unit (GRU) \citep{tang2015document}, long short-term memory networks (LSTM) \citep{hochreiter1997long}, and bi-directional long short-term memory (Bi-LSTM) \citep{schuster1997bidirectional}. Their last hidden states are used for classification. 4) Self-Attention Bi-LSTM: we apply a Bi-LSTM model with self-attention mechanism \citep{lin2017structured} and the output sentence embedding is used for classification.

We also compare our proposed model {\ModelName} with different zero-shot learning strategies:
1) DeViSE \citep{frome2013devise} finds the most compatible emerging intent label for an utterance by learning a linear compatibility function between utterances and intents; 
2) CMT \citep{socher2013zero} introduces non-linearity in the compatibility function; CMT and DeViSE are originally designed for zero-shot image classification based on pretrained CNN features. We use LSTM to encode the utterance and adopt their zero-shot learning strategies in our task;
3) CDSSM \citep{chen2016zero} uses CNN to extract character-level sentence features, where the utterance encoder shares the weights with the label encoder; 
4) Zero-shot DNN \citep{kumar2017zero} further improves the performance of CDSSM by using separate encoders for utterances and intent.
The proposed model {\ModelName} can be seen as a hybrid model: it has the advantages of the compatibility models to model the correlations between utterances and intents directly; it also explicitly derives intent representations for emerging intents without labeled utterances.

\begin{table}[ht!]
\centering
\resizebox{\linewidth}{!}{% <-
\begin{tabular}{l|llllll}
\hline
\textbf{Dataset}    & $D_W$ & $D_H$ & $D_A$ & $R$  &  $\sigma$ & $\alpha$ \\ \hline
SNIPS-NLU &  300   & 32   &  20 &  3 &   4   &   0.0001  \\ 
CVA &   200 &  200  & 100   &  8 &  1  &    0.01   \\ \hline
\end{tabular}
}
\vspace{0.1in}
\caption{Hyperparameter settings.}
\label{hyperparameter}
\vspace{-0.2in}
\end{table}

\begin{table*}[ht!]
\centering
\resizebox{1.0\textwidth}{!}{% <-
\begin{tabular}{l|cccc|cccc}
\hline
\multirow{2}{*}{\textbf{Model}} & \multicolumn{4}{c|}{\textbf{SNIPS-NLU} (on 2 emerging intents)} & \multicolumn{4}{c}{\textbf{CVA} (on 20 emerging intents)} \\ \cline{2-9} 
                      & Accuracy &Precision & Recall  & F1
                      & Accuracy &Precision & Recall  & F1              \\ \hline
DeViSE \cite{frome2013devise}  &0.7447	&0.7448	&0.7447	&0.7446   &0.7809	&0.8060	&0.7809	&0.7617\\
CMT \cite{socher2013zero}   &0.7396	&\textbf{0.8266}	&0.7396	&0.7206  &0.7721	&0.7728	&0.7721	&0.7445 \\
CDSSM \cite{chen2016zero} &0.7588	&0.7625	&0.7588	&0.7580 
&0.2140 &0.4072 &0.2140 &0.1667 \\
Zero-shot DNN \cite{kumar2017zero} &0.7165	&0.7330	&0.7165	&0.7116  &0.7903	&0.8240	&0.7903	&0.7774  \\\hline
{\ModelName} w/o Self-attention  &0.7587	&0.7764	&0.7588	&0.7547  &0.8103	&0.8512	 &0.8103	&0.8115\\ 
{\ModelName} w/o Bi-LSTM  &0.7619	&0.7631 &0.7619	&0.7616  &0.8366	&\textbf{0.8770} &0.8366	&0.8403\\ 
{\ModelName} w/o Regularizer  &0.7675	&0.7676	&0.7675	&0.7675  &0.8544	&0.8730	 &0.8544	&0.8553\\ 

\ModelName          &\textbf{0.7752}   &0.7762 &\textbf{0.7752}   &\textbf{0.7750} &\textbf{0.8628} &0.8751  &\textbf{0.8629}  &\textbf{0.8635}\\ \hline
\end{tabular}
}
\vspace{0.1in}
\caption{Zero-shot intention detection results using {\ModelName} on two datasets. All the metrics (Accuray, Precision, Recall and F1) are reported using the average value weighted by their support on per class.}
\label{tab::overall}
\vspace{-0.2in}
\end{table*}

\noindent\textbf{Implementation Details.}
The hyperparameters used for experiments are shown in Table \ref{hyperparameter}.
We use three fold cross-validation to choose hyperparameters.
The dimension of the prediction vector $D_P$ is 10 for both datasets. $D_I=D_W$ because we use the averaged word embeddings contained in the intent label as the intent embedding. An additional input dropout layer with a dropout keep rate 0.8 is applied to the SNIPS-NLU dataset.
In the loss function, the down-weighting coefficient $\lambda$ is 0.5, margins $m_k^ +$ and $m_k^ -$ are set to 0.9 and 0.1 for all the existing intents. The iteration number $iter$ used in the dynamic routing algorithm is 3.
Adam optimizer \citep{kingma2014adam} is used to minimize the loss.
\section{Results}
\noindent\textbf{Quantitative Evaluation.} 
The intention detection results on two datasets are reported in Table \ref{tab::seen_intents}, where the proposed capsule-based model {\IntentModelName} performs consistently better than bag-of-word classifiers using TF-IDF, as well as various neural network models designed for text classification.
These results demonstrate the novelty and effectiveness of the proposed capsule-based model {\IntentModelName} in modeling text for intent detection.

Also, we report results on zero-shot intention detection task in Table \ref{tab::overall}, where our model {\ModelName} outperforms other baselines that adopt different zero-shot learning strategies.
CMT has higher precision but low accuracy and recall on the SNIPS-NLU dataset.
CDSSM fails on CVA dataset, probabily because the character-level model is suitable for English corpus but not for CVA, which is in Chinese.

\noindent\textbf{Ablation Study.}
To study the contribution of different modules of {\ModelName} for zero-shot intent detection, we also report ablation test results in Table \ref{tab::overall}. ``w/o Self-attention'' is the model without self-attention: the last forward/backward hidden states of the bi-LSTM recurrent encoder are used; ``w/o Bi-LSTM'' uses the LSTM with only a forward pass;  ``w/o Regularizer'' does not encourage discrepancies among different self-attention heads: it adopts $\alpha=0$ in the loss function. Generally, from the lower part of Table \ref{tab::overall} we can see that all modules contribute to the effectiveness of the model. On the SNIPS-NLU dataset, each of the three modules has a comparable contribution to the whole model (around 2-3\% improvement in F1 score). While on the CVA dataset, the self-attention plays the most important role, which gives the model a 5.2\% improvement in F1 score.

\begin{figure}[ht!]
    \centering
    \includegraphics[width=\linewidth]{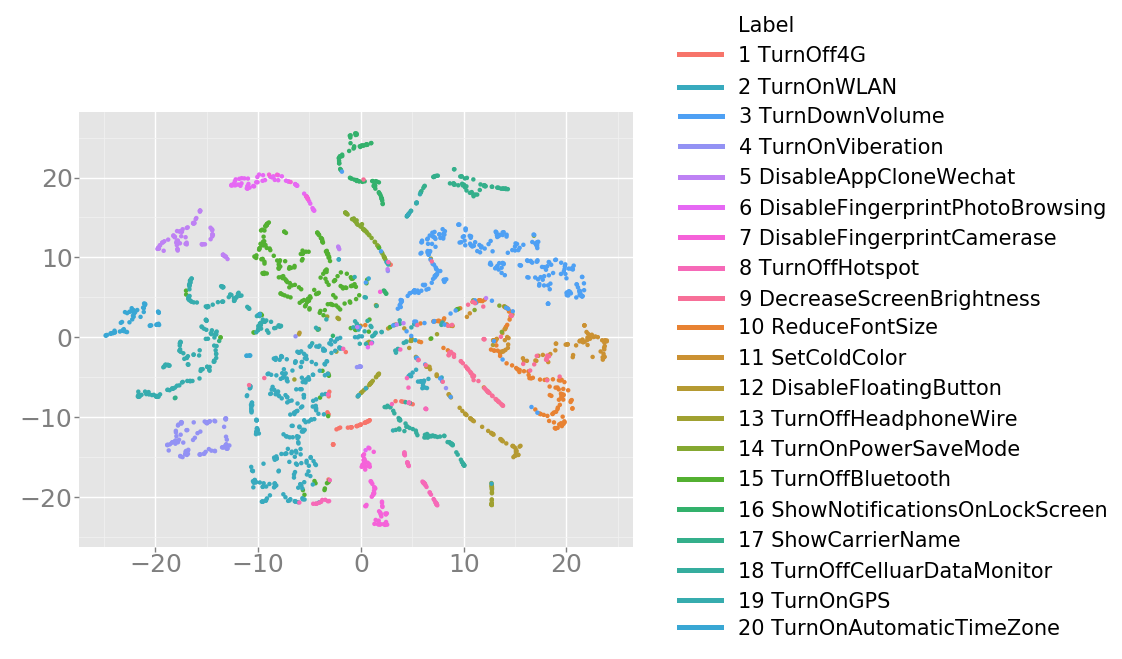}
    \vspace{0.001in}
    \caption{t-SNE visualization of normal activation vectors of utterances with 20 emerging intents in CVA.}
    \label{fig::votes_visual_huawei}
\end{figure}

\noindent\textbf{Discriminative Emerging Intent Representations.}
Besides quantitative evidences supporting the effectiveness of the {\ModelName}, we visualize activation vectors of emerging intents in Figure \ref{fig::votes_visual_huawei}.
Since the activation vectors of utterances with emerging intents are of high dimension and we are interested in their orientations which indicate their intent properties, t-SNE is applied on the normal vector of the activation vectors to reduce the dimension to 2. We color the utterances according to their ground-truth emerging intent labels.

As illustrated in Figure \ref{fig::votes_visual_huawei}, {\ModelName} has the ability to learn discriminative intent representations for emerging intents in zero-shot DetectionCaps, so that utterances with different intents naturally have different orientations. 
In the meanwhile, utterances of the same emerging intent but with nuances in expressions result in their proximity in the t-SNE space.
However, we do observe less satisfied cases where the model mistake an emerging intent \textsf{DecreaseScreenBrightness} (No. 9) with \textsf{ReduceFontSize} (No. 10) and \textsf{SetColdColor} (No. 11). When we check activation vectors of intents in Figure \ref{fig::votes_visual_huawei} we also find that these three intents tend to have similar representations around the area (15, -5). We think it is due to their inherent similarity as these three intents all try to tune display configurations.

\section{Interpretability}
Capsule models try to bring more interpretability when compared with traditional deep neural networks. We provide case studies here toward the intepretability of the proposed model in 1) extracting meaningful semantic features and 2) transferring knowledge from existing intents to emerging intents.

\noindent\textbf{Extracting Meaningful Semantic Features.}
To show that SemanticCaps have the ability to extract meaningful semantic features from the utterance, we study the self-attention matrix $\textbf{A}$ within the SemanticCaps and visualize the attention scores of utterances on both existing and emerging intents.

\begin{table}[ht!]
\centering
\resizebox{7.7cm}{!}{%
\begin{tabular}{l}
\toprule
\textbf{Existing Intent: }\textsf{PlayMusic}\\
$\bullet$ Play Action\\
\hlcy[36.428555846214294]{\texttt{\textbf{play}}} \hlcy[16.286607086658478]{\texttt{\textbf{music}}} \hlcy[2.197594568133354]{\texttt{\textbf{by}}} \hlcy[0.016248697647824883]{\texttt{\textbf{charlie}}} \hlcy[0.016248697647824883]{\texttt{\textbf{adams}}} \hlcy[0.016248697647824883]{\texttt{\textbf{from}}} \\
\hlcy[14.542676508426666]{\texttt{\textbf{i}}} \hlcy[12.053874880075455]{\texttt{\textbf{want}}} \hlcy[13.479393720626831]{\texttt{\textbf{to}}} \hlcy[12.114007771015167]{\texttt{\textbf{hear}}} \hlcy[19.378097355365753]{\texttt{\textbf{any}}} \hlcy[13.098560273647308]{\texttt{\textbf{tune}}} \hlcy[6.896718591451645]{\texttt{\textbf{from}}} \hlcy[2.634293772280216]{\texttt{\textbf{twenties}}} \\
\hlcy[27.479472756385803]{\texttt{\textbf{open}}} \hlcy[23.9264115691185]{\texttt{\textbf{up}}} \hlcy[5.287760496139526]{\texttt{\textbf{music}}} \hlcy[3.3653177320957184]{\texttt{\textbf{on}}} \hlcy[0.01069242280209437]{\texttt{\textbf{last}}} \hlcy[0.01069242280209437]{\texttt{\textbf{fm}}} \\
$\bullet$ Musician Name\\
\hlcy[0.7320448756217957]{\texttt{\textbf{}}} \hlcy[1.796233095228672]{\texttt{\textbf{i}}} \hlcy[2.7305955067276955]{\texttt{\textbf{want}}} \hlcy[4.975110664963722]{\texttt{\textbf{to}}} \hlcy[8.06172490119934]{\texttt{\textbf{hear}}} \hlcy[8.687440305948257]{\texttt{\textbf{music}}} \hlcy[9.395097196102142]{\texttt{\textbf{by}}} \hlcy[10.499640554189682]{\texttt{\textbf{madeleine}}} \hlcy[15.409015119075775]{\texttt{\textbf{peyroux}}} \hlcy[17.015331983566284]{\texttt{\textbf{from}}} \hlcy[18.71657222509384]{\texttt{\textbf{on}}} \hlcy[0.14151370851323009]{\texttt{\textbf{youtube}}} \\
\hlcy[2.727731131017208]{\texttt{\textbf{}}} \hlcy[3.258012980222702]{\texttt{\textbf{play}}} \hlcy[5.251998454332352]{\texttt{\textbf{me}}} \hlcy[8.898032456636429]{\texttt{\textbf{a}}} \hlcy[8.214519172906876]{\texttt{\textbf{song}}} \hlcy[10.193081200122833]{\texttt{\textbf{by}}} \hlcy[13.617883622646332]{\texttt{\textbf{charles}}} \hlcy[20.650646090507507]{\texttt{\textbf{neidich}}} \\
\hlcy[0.4954453557729721]{\texttt{\textbf{}}} \hlcy[2.013211138546467]{\texttt{\textbf{use}}} \hlcy[4.686387628316879]{\texttt{\textbf{itunes}}} \hlcy[7.0821113884449005]{\texttt{\textbf{to}}} \hlcy[9.928279370069504]{\texttt{\textbf{play}}} \hlcy[12.6177117228508]{\texttt{\textbf{artist}}} \hlcy[16.435439884662628]{\texttt{\textbf{ringo}}} \hlcy[18.633341789245605]{\texttt{\textbf{shiina}}} \hlcy[26.176923513412476]{\texttt{\textbf{track}}} \hlcy[0.12069696094840765]{\texttt{\textbf{in}}} \hlcy[0.12069696094840765]{\texttt{\textbf{heaven}}} \\\hline

\textbf{Existing Intent: }\textsf{SearchCreativeWork}\\
$\bullet$ Search Action\\
\hlcy[25.022611021995544]{\texttt{\textbf{find}}} \hlcy[1.1873651295900345]{\texttt{\textbf{fields}}} \hlcy[0.028829975053668022]{\texttt{\textbf{of}}} \hlcy[0.028829975053668022]{\texttt{\textbf{sacrifice}}} \hlcy[0.028829975053668022]{\texttt{\textbf{movie}}} \\
\hlcy[9.091860055923462]{\texttt{\textbf{i}}} \hlcy[22.52381592988968]{\texttt{\textbf{m}}} \hlcy[13.489031791687012]{\texttt{\textbf{looking}}} \hlcy[16.76313579082489]{\texttt{\textbf{for}}} \hlcy[12.71178275346756]{\texttt{\textbf{music}}} \hlcy[7.084269821643829]{\texttt{\textbf{of}}} \hlcy[2.1338628605008125]{\texttt{\textbf{nashville}}} \hlcy[0.005680128742824309]{\texttt{\textbf{season}}} \hlcy[0.005680128742824309]{\texttt{\textbf{saga}}} \\
\hlcy[10.171926021575928]{\texttt{\textbf{show}}} \hlcy[0.7768150884658098]{\texttt{\textbf{me}}} \hlcy[0.03563051577657461]{\texttt{\textbf{television}}} \hlcy[0.03563051577657461]{\texttt{\textbf{show}}} \hlcy[0.03563051577657461]{\texttt{\textbf{children}}} \hlcy[0.03563051577657461]{\texttt{\textbf{in}}} \hlcy[0.03563051577657461]{\texttt{\textbf{need}}} \hlcy[0.03563051577657461]{\texttt{\textbf{rocks}}} \\
$\bullet$ Creative Work Type \\
\hlcy[2.906699851155281]{\texttt{\textbf{please}}} \hlcy[4.543013870716095]{\texttt{\textbf{find}}} \hlcy[6.118370592594147]{\texttt{\textbf{me}}} \hlcy[5.793781951069832]{\texttt{\textbf{platinum}}} \hlcy[12.135510891675949]{\texttt{\textbf{box}}} \hlcy[19.604317843914032]{\texttt{\textbf{ii}}} \hlcy[24.967022240161896]{\texttt{\textbf{song}}} \hlcy[21.66745215654373]{\texttt{\textbf{?}}} \\
\hlcy[12.207721173763275]{\texttt{\textbf{show}}} \hlcy[14.69714492559433]{\texttt{\textbf{me}}} \hlcy[19.069798290729523]{\texttt{\textbf{a}}} \hlcy[25.20745098590851]{\texttt{\textbf{picture}}} \hlcy[6.137485429644585]{\texttt{\textbf{called}}} \hlcy[0.28406926430761814]{\texttt{\textbf{heart}}} \hlcy[0.28406926430761814]{\texttt{\textbf{like}}} \hlcy[0.28406926430761814]{\texttt{\textbf{a}}} \hlcy[0.28406926430761814]{\texttt{\textbf{hurricane}}} \\
\hlcy[5.238913372159004]{\texttt{\textbf{where}}} \hlcy[10.995450615882874]{\texttt{\textbf{can}}} \hlcy[11.410915851593018]{\texttt{\textbf{i}}} \hlcy[29.865694046020508]{\texttt{\textbf{buy}}} \hlcy[17.03270673751831]{\texttt{\textbf{a}}} \hlcy[22.762805223464966]{\texttt{\textbf{photograph}}} \hlcy[0.0638441473711282]{\texttt{\textbf{called}}} \hlcy[0.0638441473711282]{\texttt{\textbf{feel}}} \hlcy[0.0638441473711282]{\texttt{\textbf{love}}} \hlcy[0.0638441473711282]{\texttt{\textbf{?}}}\\
\hline
\end{tabular}%
}
\vspace{0.1in}
\caption{Attentions on utterances with existing intents on SNIPS-NLU.}\label{tab::heatmaps_seen}
\vspace{-0.2in}
\end{table}

From Table \ref{tab::heatmaps_seen} we can see that each self-attention head almost always focuses on one unique semantic feature of the utterance. For example, in the intent of \textsf{PlayMusic} one self-attention head 
always focuses on the ``play'' action while another attention focuses on musician names.
We also observe that the learned attention adopts well to diverse expressions. For example, the self-attention head in \textsf{PlayMusic} is attentive to various mentions of musician names when they are followed by words like \texttt{by},  \texttt{play} and  \texttt{artist}, even when named entities are not tagged and given to the model. The self-attention head that extracts the ``search'' action in \textsf{SearchCreativeWork} is able to be attentive to various expressions such as  \texttt{find}, \texttt{looking for} and \texttt{show}.

\noindent\textbf{Extraction-behavior Transfer by SemanticCaps.}
More importantly, we observe appealing extraction behaviors of SemanticCaps on utterances of emerging intents as well, even if they are not trained to perform semantic extraction on utterances of emerging intents.
\vspace{-0.1in}
\begin{table}[h]
\centering
\resizebox{7.7cm}{!}{%
\begin{tabular}{l}
\toprule
\textbf{Emerging Intent:} \textsf{RateBook}\\
$\bullet$ Rate Action\\
\hlcr[26.231738924980164]{\texttt{\textbf{i}}} \hlcr[28.16615104675293]{\texttt{\textbf{d}}} \hlcr[18.957585096359253]{\texttt{\textbf{rate}}} \hlcr[22.14779257774353]{\texttt{\textbf{this}}} \hlcr[3.675764799118042]{\texttt{\textbf{novel}}} \hlcr[0.45778555795550346]{\texttt{\textbf{a}}} \hlcr[0.019114512542728335]{\texttt{\textbf{five}}} \\
\hlcr[26.24477744102478]{\texttt{\textbf{add}}} \hlcr[22.31103777885437]{\texttt{\textbf{the}}} \hlcr[19.668768346309662]{\texttt{\textbf{rating}}} \hlcr[24.304300546646118]{\texttt{\textbf{for}}} \hlcr[5.62787652015686]{\texttt{\textbf{this}}} \hlcr[1.5353926457464695]{\texttt{\textbf{current}}} \hlcr[0.01620262482902035]{\texttt{\textbf{series}}} \hlcr[0.01620262482902035]{\texttt{\textbf{a}}} \hlcr[0.01620262482902035]{\texttt{\textbf{four}}} \hlcr[0.01620262482902035]{\texttt{\textbf{out}}} \hlcr[0.01620262482902035]{\texttt{\textbf{of}}} \hlcr[0.01620262482902035]{\texttt{\textbf{points}}} \\
\hlcr[49.585556983947754]{\texttt{\textbf{i}}} \hlcr[34.186601638793945]{\texttt{\textbf{give}}} \hlcr[12.639577686786652]{\texttt{\textbf{ruled}}} \hlcr[2.2200504317879677]{\texttt{\textbf{britannia}}} \hlcr[0.6442367564886808]{\texttt{\textbf{a}}} \hlcr[0.21383303683251143]{\texttt{\textbf{rating}}} \hlcr[0.026849727146327496]{\texttt{\textbf{of}}} \hlcr[0.026849727146327496]{\texttt{\textbf{five}}} \hlcr[0.026849727146327496]{\texttt{\textbf{out}}} \hlcr[0.026849727146327496]{\texttt{\textbf{of}}} \\
$\bullet$ Book Name\\
\hlcr[12.217818945646286]{\texttt{\textbf{give}}} \hlcr[15.55713266134262]{\texttt{\textbf{the}}} \hlcr[14.228272438049316]{\texttt{\textbf{televised}}} \hlcr[14.592444896697998]{\texttt{\textbf{morality}}} \hlcr[22.39955961704254]{\texttt{\textbf{series}}} \hlcr[15.589748322963715]{\texttt{\textbf{a}}} \hlcr[4.127286747097969]{\texttt{\textbf{one}}} \\
\hlcr[4.68684546649456]{\texttt{\textbf{i}}} \hlcr[6.582051515579224]{\texttt{\textbf{want}}} \hlcr[8.008881658315659]{\texttt{\textbf{to}}} \hlcr[9.726732224225998]{\texttt{\textbf{give}}} \hlcr[11.91072165966034]{\texttt{\textbf{the}}} \hlcr[9.469421952962875]{\texttt{\textbf{coming}}} \hlcr[13.528846204280853]{\texttt{\textbf{of}}} \hlcr[13.675844669342041]{\texttt{\textbf{the}}} \hlcr[12.004149705171585]{\texttt{\textbf{terraphiles}}} \hlcr[8.014107495546341]{\texttt{\textbf{a}}} \hlcr[2.314702421426773]{\texttt{\textbf{rating}}} \hlcr[0.005549772686208598]{\texttt{\textbf{of}}} \\
\hlcr[23.371325433254242]{\texttt{\textbf{the}}} \hlcr[23.911534249782562]{\texttt{\textbf{chronicle}}} \hlcr[23.79162460565567]{\texttt{\textbf{charlie}}} \hlcr[15.976467728614807]{\texttt{\textbf{peace}}} \hlcr[12.067839503288269]{\texttt{\textbf{earns}}} \hlcr[0.5490751005709171]{\texttt{\textbf{stars}}} \hlcr[0.08944115252234042]{\texttt{\textbf{from}}} \hlcr[0.013482285430654883]{\texttt{\textbf{me}}} \\
$\bullet$ Rating Score\\
\hlcr[3.6312922835350037]{\texttt{\textbf{rate}}} \hlcr[4.353851079940796]{\texttt{\textbf{the}}} \hlcr[6.112312898039818]{\texttt{\textbf{grisly}}} \hlcr[6.33174404501915]{\texttt{\textbf{wife}}} \hlcr[11.307782679796219]{\texttt{\textbf{three}}} \hlcr[13.766269385814667]{\texttt{\textbf{points}}} \hlcr[18.268045783042908]{\texttt{\textbf{out}}} \hlcr[16.32569432258606]{\texttt{\textbf{of}}} \hlcr[16.32569432258606]{\texttt{\textbf{five}}} \\
\hlcr[2.9697757214307785]{\texttt{\textbf{i}}} \hlcr[3.9993945509195328]{\texttt{\textbf{would}}} \hlcr[4.8072244971990585]{\texttt{\textbf{give}}} \hlcr[8.336102217435837]{\texttt{\textbf{this}}} \hlcr[7.396101951599121]{\texttt{\textbf{current}}} \hlcr[8.56083482503891]{\texttt{\textbf{chronicle}}} \hlcr[25.208842754364014]{\texttt{\textbf{three}}} \hlcr[29.72981631755829]{\texttt{\textbf{points}}} \\
\hlcr[2.5003425776958466]{\texttt{\textbf{this}}} \hlcr[5.52409403026104]{\texttt{\textbf{saga}}} \hlcr[5.88969811797142]{\texttt{\textbf{deserves}}} \hlcr[5.524230748414993]{\texttt{\textbf{a}}} \hlcr[13.049910962581635]{\texttt{\textbf{score}}} \hlcr[15.115945041179657]{\texttt{\textbf{of}}} \hlcr[20.587673783302307]{\texttt{\textbf{four}}} \\\hline

\textbf{Emerging Intent:} \textsf{AddToPlaylist}\\
$\bullet$ Song/Artist Name \\
\hlcr[12.947624921798706]{\texttt{\textbf{add}}} \hlcr[17.263276875019073]{\texttt{\textbf{star}}} \hlcr[15.372216701507568]{\texttt{\textbf{light}}} \hlcr[14.67978060245514]{\texttt{\textbf{star}}} \hlcr[17.50292181968689]{\texttt{\textbf{bright}}} \hlcr[15.721015632152557]{\texttt{\textbf{to}}} \hlcr[5.048931390047073]{\texttt{\textbf{my}}} \hlcr[1.312430016696453]{\texttt{\textbf{jazz}}} \hlcr[0.008929320028983057]{\texttt{\textbf{classics}}} \hlcr[0.008929320028983057]{\texttt{\textbf{playlist}}} \\
\hlcr[8.010569214820862]{\texttt{\textbf{i}}} \hlcr[10.519406199455261]{\texttt{\textbf{want}}} \hlcr[10.338525474071503]{\texttt{\textbf{a}}} \hlcr[9.50050950050354]{\texttt{\textbf{song}}} \hlcr[14.559109508991241]{\texttt{\textbf{by}}} \hlcr[13.705258071422577]{\texttt{\textbf{john}}} \hlcr[13.287101686000824]{\texttt{\textbf{schlitt}}} \hlcr[11.029616743326187]{\texttt{\textbf{in}}} \hlcr[6.982363015413284]{\texttt{\textbf{the}}} \hlcr[1.9776221364736557]{\texttt{\textbf{bajo}}} \hlcr[0.005994487946736626]{\texttt{\textbf{las}}} \hlcr[0.005994487946736626]{\texttt{\textbf{estrellas}}} \hlcr[0.005994487946736626]{\texttt{\textbf{playlist}}} \\
\hlcr[24.894385039806366]{\texttt{\textbf{put}}} \hlcr[30.802133679389954]{\texttt{\textbf{sungmin}}} \hlcr[20.72480469942093]{\texttt{\textbf{into}}} \hlcr[20.971810817718506]{\texttt{\textbf{my}}} \hlcr[1.6725504770874977]{\texttt{\textbf{summer}}} \hlcr[0.6230746395885944]{\texttt{\textbf{playlist}}} \\
$\bullet$ Playlist Name\\
\hlcr[1.9873594865202904]{\texttt{\textbf{add}}} \hlcr[3.1764842569828033]{\texttt{\textbf{an}}} \hlcr[5.157268047332764]{\texttt{\textbf{album}}} \hlcr[5.7072702795267105]{\texttt{\textbf{to}}} \hlcr[6.437981128692627]{\texttt{\textbf{my}}} \hlcr[6.989925354719162]{\texttt{\textbf{list}}} \hlcr[24.44448173046112]{\texttt{\textbf{la}}} \hlcr[36.64641082286835]{\texttt{\textbf{mejor}}} \hlcr[0.5560481920838356]{\texttt{\textbf{música}}} \hlcr[0.5560481920838356]{\texttt{\textbf{dance}}} \\
\hlcr[1.7827222123742104]{\texttt{\textbf{can}}} \hlcr[3.197145089507103]{\texttt{\textbf{you}}} \hlcr[3.506917506456375]{\texttt{\textbf{add}}} \hlcr[3.5877488553524017]{\texttt{\textbf{danny}}} \hlcr[7.619887590408325]{\texttt{\textbf{carey}}} \hlcr[14.48204517364502]{\texttt{\textbf{to}}} \hlcr[21.563702821731567]{\texttt{\textbf{my}}} \hlcr[37.507474422454834]{\texttt{\textbf{masters}}} \hlcr[0.3971970174461603]{\texttt{\textbf{of}}} \hlcr[0.3971970174461603]{\texttt{\textbf{metal}}} \hlcr[0.3971970174461603]{\texttt{\textbf{playlist}}} \\
\hlcr[1.1785421520471573]{\texttt{\textbf{i}}} \hlcr[0.9884560480713844]{\texttt{\textbf{want}}} \hlcr[1.92047581076622]{\texttt{\textbf{to}}} \hlcr[1.7881151288747787]{\texttt{\textbf{put}}} \hlcr[3.3012904226779938]{\texttt{\textbf{a}}} \hlcr[6.678289920091629]{\texttt{\textbf{copy}}} \hlcr[6.223602965474129]{\texttt{\textbf{of}}} \hlcr[5.464228987693787]{\texttt{\textbf{this}}} \hlcr[11.14695742726326]{\texttt{\textbf{tune}}} \hlcr[10.771258175373077]{\texttt{\textbf{into}}} \hlcr[10.823307186365128]{\texttt{\textbf{skatepark}}} \hlcr[17.095008492469788]{\texttt{\textbf{punks}}} \\\hline
\end{tabular}%
}
\vspace{0.1in}
\caption{Attentions on utterances with emerging intents on SNIPS-NLU.}\label{tab::heatmaps_unseen}
\end{table}

From Table \ref{tab::heatmaps_unseen} we observe that the same self-attention head that extracts ``play'' action in the existing intent \textsf{PlayMusic} is also attentive to  words or phrases referring to the ``rate'' action in an emerging intent \textsf{RateABook}: like \texttt{rate}, \texttt{add the rating}, and \texttt{give}. Other self-attention heads are almost always focusing on other aspects of the utterances such as the book name or the actual rating score.

Such behavior not only shows that SemanticCaps have the capacity to learn an intent-independent semantic feature extractor, which extracts generalizable semantic features that either existing or emerging intent representations are built upon, but also indicates that SemanticCaps has the ability to transfer extraction behaviors among utterances of different intents.

\begin{figure}[h!]
    \centering
    \includegraphics[width=\linewidth]{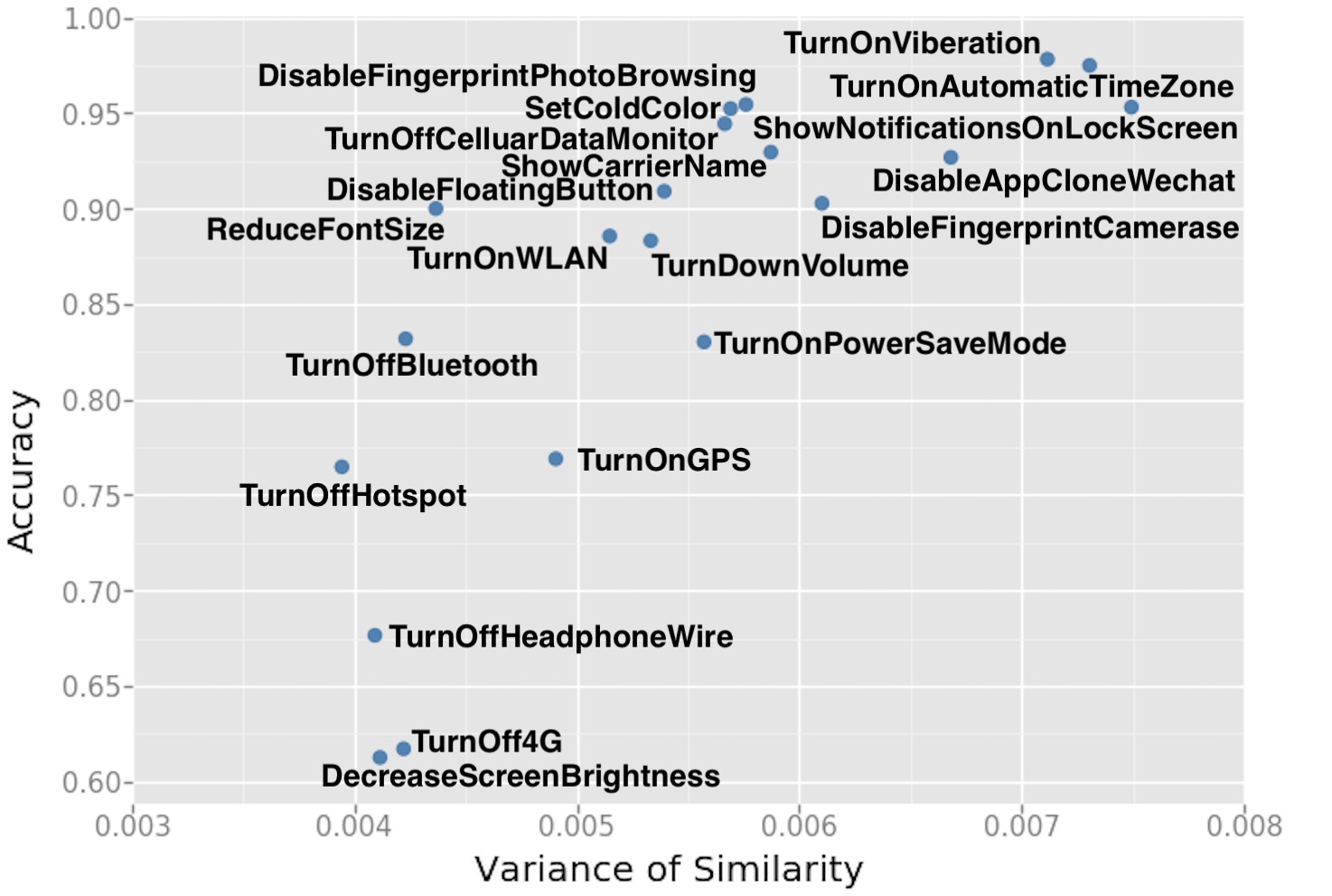}
    \vspace{0.1in}
    \caption{Accuracy vs. variance of the similarity distribution for 20 emerging intents in CVA dataset.}
    \label{fig:similarity}
\end{figure}

\noindent\textbf{Knowledge Transfer via Intent Similarity.}
Beside extracting semantic features and utilizing existing routing information,
we use similarities between intent embeddings to help transfer vote vectors from {\IntentModelName} to {\ModelName}. 
We study the similarity distribution of each emerging intents to all existing intents in Figure \ref{fig:similarity}.

The y axis is the zero-shot detection accuracy on each emerging intent in the CVA dataset. The x axis measures $\text{var}(\mathbf{q}_l)$, the variance of the similarity distribution of each emerging intent $l$ to all the existing intents.
If an emerging intent has a high variance in the similarity distribution, it means that some existing intents have higher similarities with this emerging intent than others: the model is more certain about which existing intent to transfer the similarity knowledge from, based on intent label similarities.
In this case, 13 out of 20 emerging intents with high variances where $\text{var}(\mathbf{q}_l)>0.005$ always have a decent performance (Accuracy$\geqslant$0.83).
While a low variance does not necessarily always lead to less satisfied performances as some intents can rely on existing intents more evenly together, but with less confidence on each, for knowledge transfer.
\section{Conclusions}
In this paper, a capsule-based model, namely {\IntentModelName}, is first introduced to harness the advantages of capsule models for text modeling in a hierarchical manner: semantic features are extracted from the utterances with self-attention, and aggregated via the dynamic routing-by-agreement mechanism to obtain utterance-level intent representations. We believe that the inductive biases subsumed in such capsule-based hierarchical learning schema have broader applicability on various text modeling tasks, besides its evidenced performance on the intent detection task we studied in this paper.
The proposed {\ModelName} model further introduces zero-shot learning ability to the capsule model via various means of knowledge transfer from existing intents for discriminating emerging intents where no labeled utterances or excessive external resources are available. 
Experiments on two real-world datasets show the effectiveness and intepretability of the proposed models.
\section{Acknowledgments}
We thank the reviewers for their valuable comments.
This work is supported in part by NSF through grants IIS-1526499, IIS-1763325, and CNS-1626432, and NSFC 61672313.
Xiaohui Yan's work is funded by the National Natural Science Foundation of China (NSFC) under Grant No. 61502447.

\balance
\bibliography{ref.bib}

\begin{thebibliography}{27}
\expandafter\ifx\csname natexlab\endcsname\relax\def\natexlab#1{#1}\fi

\bibitem[{Changpinyo et~al.(2016)Changpinyo, Chao, Gong, and
  Sha}]{changpinyo2016synthesized}
Soravit Changpinyo, Wei-Lun Chao, Boqing Gong, and Fei Sha. 2016.
\newblock Synthesized classifiers for zero-shot learning.
\newblock In \emph{CVPR}, pages 5327--5336.

\bibitem[{Chen et~al.(2016{\natexlab{a}})Chen, Hakkani-T{\"u}r, and
  He}]{chen2016zero}
Yun-Nung Chen, Dilek Hakkani-T{\"u}r, and Xiaodong He. 2016{\natexlab{a}}.
\newblock Zero-shot learning of intent embeddings for expansion by
  convolutional deep structured semantic models.
\newblock In \emph{ICASSP}, pages 6045--6049.

\bibitem[{Chen et~al.(2016{\natexlab{b}})Chen, Hakkani-T{\"u}r, T{\"u}r, Gao,
  and Deng}]{chen2016end}
Yun-Nung Chen, Dilek Hakkani-T{\"u}r, G{\"o}khan T{\"u}r, Jianfeng Gao, and
  Li~Deng. 2016{\natexlab{b}}.
\newblock End-to-end memory networks with knowledge carryover for multi-turn
  spoken language understanding.
\newblock In \emph{INTERSPEECH}, pages 3245--3249.

\bibitem[{Ferreira et~al.(2015{\natexlab{a}})Ferreira, Jabaian, and
  Lefevre}]{ferreira2015online}
Emmanuel Ferreira, Bassam Jabaian, and Fabrice Lefevre. 2015{\natexlab{a}}.
\newblock Online adaptative zero-shot learning spoken language understanding
  using word-embedding.
\newblock In \emph{ICASSP}, pages 5321--5325.

\bibitem[{Ferreira et~al.(2015{\natexlab{b}})Ferreira, Jabaian, and
  Lef{\`e}vre}]{ferreira2015zero}
Emmanuel Ferreira, Bassam Jabaian, and Fabrice Lef{\`e}vre. 2015{\natexlab{b}}.
\newblock Zero-shot semantic parser for spoken language understanding.
\newblock In \emph{INTERSPEECH}, pages 1403--1407.

\bibitem[{Frome et~al.(2013)Frome, Corrado, Shlens, Bengio, Dean, Mikolov
  et~al.}]{frome2013devise}
Andrea Frome, Greg~S Corrado, Jon Shlens, Samy Bengio, Jeff Dean, Tomas
  Mikolov, et~al. 2013.
\newblock Devise: A deep visual-semantic embedding model.
\newblock In \emph{NIPS}, pages 2121--2129.

\bibitem[{Hinton et~al.(2011)Hinton, Krizhevsky, and
  Wang}]{hinton2011transforming}
Geoffrey~E Hinton, Alex Krizhevsky, and Sida~D Wang. 2011.
\newblock Transforming auto-encoders.
\newblock In \emph{ICANN}, pages 44--51.

\bibitem[{Hochreiter and Schmidhuber(1997)}]{hochreiter1997long}
Sepp Hochreiter and J{\"u}rgen Schmidhuber. 1997.
\newblock Long short-term memory.
\newblock \emph{Neural computation}, 9(8):1735--1780.

\bibitem[{Hoy(2018)}]{hoy2018alexa}
Matthew~B Hoy. 2018.
\newblock Alexa, siri, cortana, and more: An introduction to voice assistants.
\newblock \emph{Medical reference services quarterly}, 37(1):81--88.

\bibitem[{Hu et~al.(2009)Hu, Wang, Lochovsky, Sun, and
  Chen}]{hu2009understanding}
Jian Hu, Gang Wang, Fred Lochovsky, Jian-tao Sun, and Zheng Chen. 2009.
\newblock Understanding user's query intent with wikipedia.
\newblock In \emph{WWW}, pages 471--480.

\bibitem[{Kim(2014)}]{kim2014convolutional}
Yoon Kim. 2014.
\newblock Convolutional neural networks for sentence classification.
\newblock \emph{arXiv preprint arXiv:1408.5882}.

\bibitem[{Kingma and Ba(2014)}]{kingma2014adam}
Diederik~P Kingma and Jimmy Ba. 2014.
\newblock Adam: A method for stochastic optimization.
\newblock \emph{arXiv preprint arXiv:1412.6980}.

\bibitem[{Kumar et~al.(2017)Kumar, Muddireddy, Dreyer, and
  Hoffmeister}]{kumar2017zero}
Anjishnu Kumar, Pavankumar~Reddy Muddireddy, Markus Dreyer, and Bj{\"o}rn
  Hoffmeister. 2017.
\newblock Zero-shot learning across heterogeneous overlapping domains.
\newblock In \emph{INTERSPEECH}, volume 2017, pages 2914--2918.

\bibitem[{Lafferty et~al.(2001)Lafferty, McCallum, and
  Pereira}]{lafferty2001conditional}
John~D. Lafferty, Andrew McCallum, and Fernando C.~N. Pereira. 2001.
\newblock Conditional random fields: Probabilistic models for segmenting and
  labeling sequence data.
\newblock In \emph{ICML}, pages 282--289.

\bibitem[{Lampert et~al.(2014)Lampert, Nickisch, and
  Harmeling}]{lampert2014attribute}
Christoph~H Lampert, Hannes Nickisch, and Stefan Harmeling. 2014.
\newblock Attribute-based classification for zero-shot visual object
  categorization.
\newblock \emph{IEEE Transactions on Pattern Analysis and Machine
  Intelligence}, 36(3):453--465.

\bibitem[{Lin et~al.(2017)Lin, Feng, Santos, Yu, Xiang, Zhou, and
  Bengio}]{lin2017structured}
Zhouhan Lin, Minwei Feng, Cicero Nogueira~dos Santos, Mo~Yu, Bing Xiang, Bowen
  Zhou, and Yoshua Bengio. 2017.
\newblock A structured self-attentive sentence embedding.
\newblock In \emph{ICLR}.

\bibitem[{Liu and Lane(2016)}]{liu2016attention}
Bing Liu and Ian Lane. 2016.
\newblock Attention-based recurrent neural network models for joint intent
  detection and slot filling.
\newblock In \emph{INTERSPEECH}, pages 685--689.

\bibitem[{Mikolov et~al.(2013)Mikolov, Chen, Corrado, and
  Dean}]{mikolov2013efficient}
Tomas Mikolov, Kai Chen, Greg Corrado, and Jeffrey Dean. 2013.
\newblock Efficient estimation of word representations in vector space.
\newblock \emph{arXiv preprint arXiv:1301.3781}.

\bibitem[{Sabour et~al.(2017)Sabour, Frosst, and Hinton}]{sabour2017dynamic}
Sara Sabour, Nicholas Frosst, and Geoffrey~E Hinton. 2017.
\newblock Dynamic routing between capsules.
\newblock In \emph{NIPS}, pages 3859--3869.

\bibitem[{Schuster and Paliwal(1997)}]{schuster1997bidirectional}
Mike Schuster and Kuldip~K Paliwal. 1997.
\newblock Bidirectional recurrent neural networks.
\newblock \emph{IEEE Transactions on Signal Processing}, 45(11):2673--2681.

\bibitem[{Socher et~al.(2013)Socher, Ganjoo, Manning, and Ng}]{socher2013zero}
Richard Socher, Milind Ganjoo, Christopher~D Manning, and Andrew Ng. 2013.
\newblock Zero-shot learning through cross-modal transfer.
\newblock In \emph{NIPS}, pages 935--943.

\bibitem[{Tang et~al.(2015)Tang, Qin, and Liu}]{tang2015document}
Duyu Tang, Bing Qin, and Ting Liu. 2015.
\newblock Document modeling with gated recurrent neural network for sentiment
  classification.
\newblock In \emph{EMNLP}, pages 1422--1432.

\bibitem[{Vaswani et~al.(2017)Vaswani, Shazeer, Parmar, Uszkoreit, Jones,
  Gomez, Kaiser, and Polosukhin}]{vaswani2017attention}
Ashish Vaswani, Noam Shazeer, Niki Parmar, Jakob Uszkoreit, Llion Jones,
  Aidan~N Gomez, {\L}ukasz Kaiser, and Illia Polosukhin. 2017.
\newblock Attention is all you need.
\newblock In \emph{NIPS}, pages 6000--6010.

\bibitem[{{Watson Assistant}(2017)}]{ibm2017doc}
IBM {Watson Assistant}. 2017.
\newblock Defining intents.
\newblock In
  \emph{\url{https://console.bluemix.net/docs/services/conversation/intents.html\#defining-intents}}.

\bibitem[{Xu and Sarikaya(2013)}]{xu2013convolutional}
Puyang Xu and Ruhi Sarikaya. 2013.
\newblock Convolutional neural network based triangular crf for joint intent
  detection and slot filling.
\newblock In \emph{ASRU}, pages 78--83.

\bibitem[{Yazdani and Henderson(2015)}]{yazdani2015model}
Majid Yazdani and James Henderson. 2015.
\newblock A model of zero-shot learning of spoken language understanding.
\newblock In \emph{EMNLP}, pages 244--249.

\bibitem[{Zhang et~al.(2016)Zhang, Fan, Du, and Yu}]{zhang2016mining}
Chenwei Zhang, Wei Fan, Nan Du, and Philip~S Yu. 2016.
\newblock Mining user intentions from medical queries: A neural network based
  heterogeneous jointly modeling approach.
\newblock In \emph{WWW}, pages 1373--1384.

\end{thebibliography}
\bibliographystyle{acl_natbib_nourl}
\end{document}